\documentclass[letterpaper, 10 pt, conference]{ieeeconf}  
\usepackage{times}  
\usepackage{helvet} 
\usepackage{courier}  
\usepackage[hyphens]{url}  
\usepackage{graphicx} 
\usepackage{graphicx}  
\usepackage{graphicx}
\usepackage{amsmath}

\usepackage{hyperref}
\usepackage{url}
\usepackage{graphicx}
\usepackage{amsmath,amsfonts,amssymb,mathrsfs}
\usepackage{graphicx}
\usepackage{psfrag,graphicx,epsfig,epsf}
\usepackage[ruled,linesnumbered,vlined]{algorithm2e}
\usepackage{fancyhdr}
\usepackage{wrapfig}
\usepackage{subfigure}
\usepackage{mathtools}
\usepackage{url}
\usepackage{color}
\usepackage{verbatim}
\usepackage{xspace}

\usepackage{algpseudocode}
\usepackage{cite}

\IEEEoverridecommandlockouts                              

\title{\LARGE \bf
Learning Transition Models with Time-delayed Causal Relations}

\author{
Junchi Liang  and Abdeslam Boularias$^{1}$
\thanks{$^{1}$The authors are with the Department of Computer Science of Rutgers University, Piscataway, New Jersey 08854, USA.
        {\tt\footnotesize \{jl2068, ab1544\}@cs.rutgers.edu}}%
}

\begin{document}

\maketitle
\thispagestyle{empty}
\pagestyle{empty}

\begin{abstract}

This paper introduces an algorithm for discovering implicit and delayed causal relations between events observed by a robot at  arbitrary times, with the objective of improving data-efficiency and interpretability of model-based reinforcement learning (RL) techniques. The proposed algorithm initially predicts observations with the Markov assumption, and incrementally introduces new hidden variables to explain and reduce the stochasticity of the observations. The hidden variables are memory units that keep track of pertinent past events. Such events are systematically identified by their information gains. The learned transition and reward models are then used for planning. 
Experiments on simulated and real robotic tasks show that this method significantly improves over current RL techniques. 
\end{abstract}

\section{Introduction}
The advent of deep learning made a deep impact on most areas of computing, and provided dramatically improved solutions to many real-world problems. The appeal of deep learning is due to its simplicity that requires minimal design efforts, combined with a capability to learn complex functions from data. It did not take very long for this thrust to reach and transform the area of reinforcement learning (RL). The seminal work of~\cite{mnih2013playing} has shown that a simple neural network (DQN) could be trained to play Atari video games at a human level, using only raw screen images as observations and keyboard inputs as actions. The work on DQN  paved the way for several new techniques that can be categorized under the general umbrella of {\it end-to-end visual RL}. These techniques avoid the tedious process of designing features manually, and rely on convolutional layers to automatically extract features from sensory inputs. 

Despite the remarkable progress made by deep RL agents in reaching human-level performance and beyond~\cite{mnih2013playing}, they continue to lag behind humans in terms of data efficiency. Humans can immediately figure out the effects of their actions on objects displayed on a screen after a few trials, and build a model for reasoning and planning in order to improve their scores. Model-based RL algorithms arguably require less data than model-free ones~\cite{hafner2018learning,Finn2016DeepSA,7989324}. But learning models that are sufficiently accurate for planning is still a challenging problem~\cite{DBLP:conf/nips/BattagliaPLRK16}. Inaccurate predictions generally result in sub-optimal policies. 

The difficulty in learning accurate predictive models can be mostly attributed to the partial observability of the states. 
\begin{figure}
   \centering
    \includegraphics[width=0.5\textwidth]{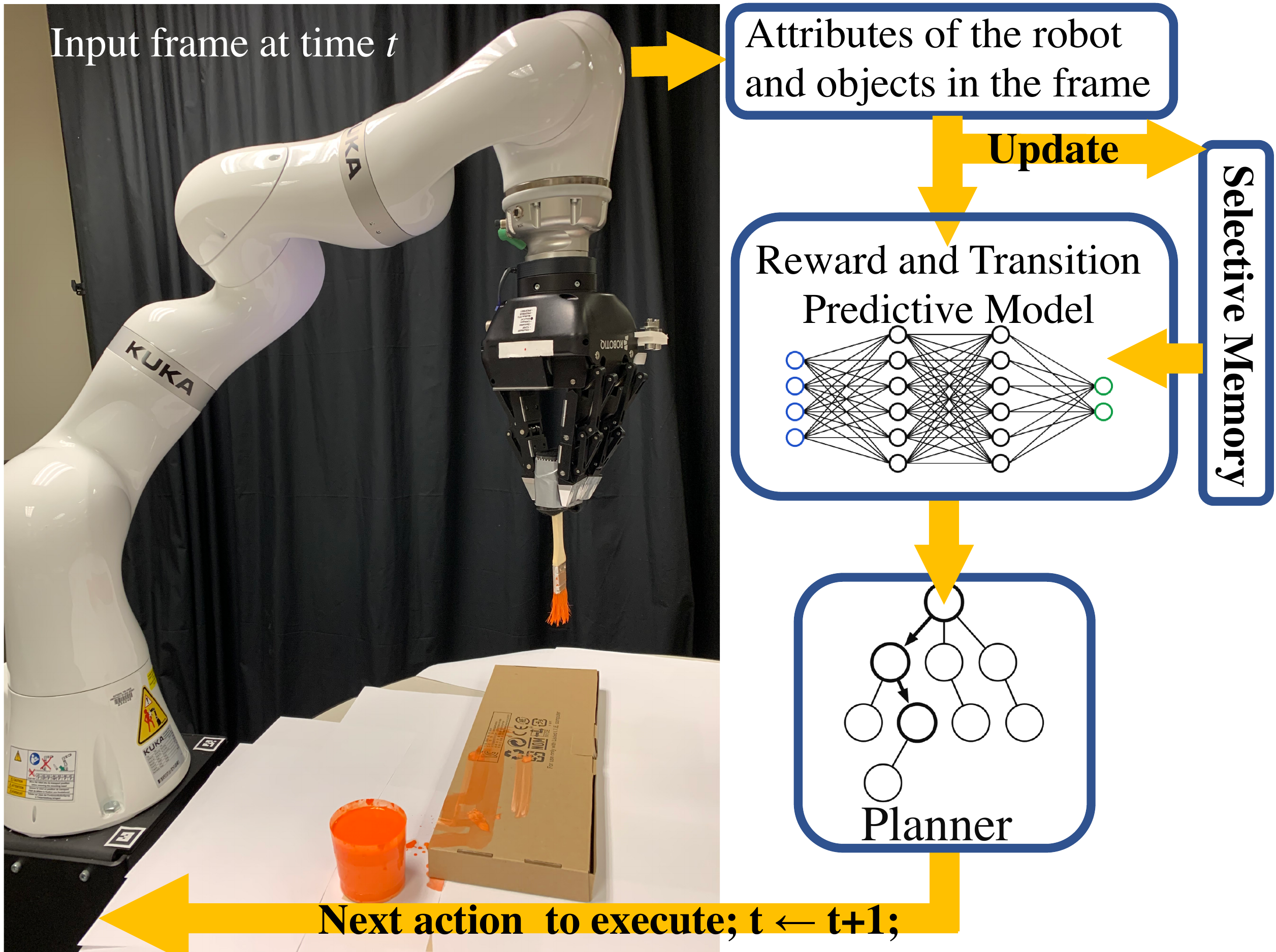}
 \caption{\small Overview of the proposed system and the robotic setup used in the experiment {\it Learning to Paint}}
 \label{fig:system}
 \vspace{-0.4cm}
 \end{figure}
In robotics, for example, the Markov condition is seldom verified. Future states and rewards often depend on the entire history of actions and observations. 

An example of that is unlocking a door and opening it afterwards. The first action changes the hidden variable that is the state of the door from {\tt locked} to {\tt unlocked}. 
Without a memory of past locking/unlocking actions, a robot cannot explain why the door opens sometimes and does not open at other times based only on an image of the door.
The robot needs to infer the hidden {\it causal link} between the act of unlocking the door and the ability to open it later in the future. 
 Other examples include filling or pouring liquids from containers, turning electric switches and unscrewing a lid and lifting it later in the future. In general,  sequential object manipulation tasks involve changing states of objects in ways that cannot be easily perceived through vision, but their effects can be observed in the later stages of the task. Figure~\ref{fig:system} shows an experiment performed in the present work, where a robot learns to paint. The causal link between dunking the paintbrush in a paint container and the appearance of paint on the surface of the box later when the brush is pressed against it is non-trivial because the robot performs a large number of random exploratory actions between the two events. The brush always looks the same, even after the paint in it has dried. Thus, it is important to remember the event of moving the brush into the container in order to predict future observations.

LSTM and GRU architectures are general-purpose tools for solving problems of partial observability by discovering and remembering pertinent information. They tend, however, to require large amounts of data, and they cannot be easily interpreted. To address these two issues, we present here an approach that combines the merits of general function approximators such as neural networks with probabilistic graphical models for representing hidden variables. 
Given a stream of actions, observations and rewards, a neural network is trained to predict future observations and rewards. Simultaneously, a graphical model of causal relations between observations occurring at different time steps is also gradually constructed. The values of the variables in the graph are also provided to the neural network as additional inputs along with the observations. The learned predictive model is then utilized by the agent to select actions based on their predicted future rewards. 

\section{Related Works}
{\bf Learning POMDPs}: The Baum-Welch algorithm is an expectation-maximization technique that is traditionally used to learn hidden Markov models and POMDPs~\cite{Rabiner:1990:THM:108235.108253}. This algorithm requires knowing the number of hidden variables in advance. It is sensitive to the initial values of parameters and typically results in suboptimal solutions. Predictive state representations (PSRs)~\cite{Singh:2004:PSR,6827} are an alternative model to represent partially observable environments without using hidden states. While parameters of PSRs can be learned with any consistent density estimator, discovering the core tests is still an open problem.
Moreover, the learning complexity of these estimators is exponential in the length of history that needs to be stacked to predict future observations. A more efficient spectral approach proposed in~\cite{Boots:2011:CLL:2000201.2000208} generalizes PSRs by including features of test outcomes and
histories, instead of a stream of raw observations. For example, an indicative feature might be the number of times we saw a specific observation in the past three steps. The memory variables introduced in the present work are closely related to the indicative features. In contrast to~\cite{Boots:2011:CLL:2000201.2000208}, our algorithm reasons about the causal relations between different regions of the observation space and over different time intervals, and returns an explicit graphical representation of the discovered causal relations. 
Most recent efforts on learning partially observable dynamical models rely on recurrent neural networks (RNN) and LSTM techniques in particular~\cite{Downey-NIPS-17,Choromanski-ICLR-18,hafner2018learning,7989324,Finn2016DeepSA}. While the success of LSTM is not fully understood, it is frequently attributed to the gating mechanisms that allows information to be retained for a long time, but also to be forgotten quickly. 

{\bf Object-oriented RL}:
In vision-based RL, there is a clear physical structure that can readily be exploited. Images can be decomposed into segments of objects. Several models utilizing object-oriented representations for learning and planning have been proposed in the past~\cite{Diuk:2008:ORE:1390156.1390187,pmlr-v32-scholz14,DBLP:journals/corr/UsunierSLC16}. Object-sensitive deep RL is a closely related idea proposed in~\cite{GCAI2017}. More recent works focused on learning these models, such as the {\it interaction networks} \cite{DBLP:conf/nips/BattagliaPLRK16}, which can reason about how objects in complex systems interact. The {\it schema network}, proposed recently by~\cite{DBLP:conf/icml/KanskySMELLDSPG17}, is an object-oriented generative physics simulator capable of disentangling multiple causes of events that occur in visual RL; it has been shown to increase transferability of skills within variations of Atari games. The schema networks are trained with an algorithm for structure learning in graphical models, but no hidden variables or delayed cause-effect relations were considered, which are the focus of our work. Object-oriented RL has not yet been applied to 3D games, to the best of our knowledge.


{\bf Memory models}: Long-term dependencies in temporal models were considered in some recent works~\cite{neitz2018adaptive,trinh2018learning}. For example, the method presented in~\cite{neitz2018adaptive} classifies trajectories based on the probabilities of future observations after an unknown number of steps. This approach is closely related to ours, except that we look into past events instead of future trajectories. ~\cite{trinh2018learning} proposed another closely related approach based on using the reconstruction loss in recurrent neural nets as an auxiliary objective. 

{\bf Attention mechanism}: Attention mechanism is widely used for selecting specific features or specific parts of features dynamically according to the specified task. This mechanism is usually characterized by aggregating a collection of features through a weighted sum where weights are functions of inputs rather than learnable parameters. One of the successful applications of attention mechanisms is {\it caption generation for images}. A preeminent work, \cite{pmlr-v37-xuc15}, defines attention weights as functions of visual features from different parts of an input image and from states of recurrent units in order to guide the architecture to focus on pertinent regions of the image as it generates a caption. The same mechanism was adopted in a more recent work~\cite{jiang2018recurrent}. Followup works explored various designs of the attention function. For instance,~\cite{anderson2018bottom} investigates two new modules, one measures magnitudes of embedded features in attention computation while the other one takes object information as part of the input in order to obtain an attention module in a higher level.~\cite{DuanASHSSAZ17} takes advantage of attention mechanism to efficiently compress information from demonstrated trajectories in the setting of imitation learning. 
 In this work, we focus on attention on specific events in the construction of memory units. Considering the forget and input gate as attention weights, LSTM inherently employs attention mechanisms~\cite{DuanASHSSAZ17}. We show that its attention tends to forget old events, while our proposed algorithm avoids this drawback. 

\section{Background and Notations}
Formally, a Markov Decision Process (MDP) is a tuple
$( \mathcal{S},  \mathcal{A}, T, R, \gamma)$, where
$ \mathcal{S}$ is a set of states and $\mathcal{A}$ is a set
of actions. $T$ is a transition function with $T(s'|a,s)=P(S_{t+1}=s'|S_{t}=s,A_{t}=a)$ for $s,s'\in  \mathcal{S}, a \in \mathcal{A}$, and $R$ is a reward function where $ R(s,a,r)$ is the probability of receiving reward $r\in \mathbb{R}$ for executing $a$ in $s$. A policy $\pi$ is a distribution on the action
to be executed in each state, defined as $\pi(s,a)=P(a_t=a|s_t=s)$.
The value $V^{\pi}$ of a policy $\pi$ is the expected sum of rewards
that will be received if $\pi$ is followed, i.e., $ V^{\pi}(s) = \mathbb{E}
[\sum_{t=0}^{\infty} \gamma^t r_t|S_0=s,\pi,T,R] $.

In robotics, states generally cannot be fully observed. Instead, a robot perceives partial observations $z_t$, in the form of images, for example. The resulting process is a Partially Observable MDP (POMDP). 
Formally, a POMDP is a tuple $( \mathcal{S},  \mathcal{A}, \mathcal{Z}, T, F, R, \gamma)$ where $( \mathcal{S}, \mathcal{A}, T, R, \gamma)$ is an MDP, $Z$ is a set of observations, and $F$ is an observation likelihood
function: $F(Z|S)$ is the probability of observing $Z\in \mathcal{Z}$ in state $S\in \mathcal{S}$.

We focus in this work on {\it object-oriented} POMDPs, where a state $S$ is described by one or several visible attributes of objects, in addition to one or several hidden variables. 
In other terms, $S = (O^1,O^2, \dots,O^n, M^1,M^2,\dots, M^m) \in \prod_{i=1}^{n} \mathcal{O}_i \times \prod_{i=1}^{m} \mathcal{M}_i$, wherein $O^i$ is an attribute of an object in the scene, and $M^i$ is a hidden variable. The list of objects includes the robot or its end-effector. $\mathcal{O}_i$ and $\mathcal{M}_i$ denote the domains of visible variable $O^i$ and hidden variable $M^i$, respectively, which can be discrete or continuous.  For example, $O^{1}$ is the position of a specific object in the image, while $O^{2}$ is its velocity, $O^{3}$ is its size, $O^{4}$ is its numerical label, $O^{5}$ is a Boolean attribute that indicates if the object still exists in the scene or not, $O^{5}$ is the position of a second object in the image, and so forth. 
Observations correspond to object attributes, i.e., $Z_t = (O_t^i)_{i=1}^{n}$. To ease the notation, we also consider the reward signal $R_t$ as one of the observable attributes $O_t^i$. Thus, the reward function is modeled and learned in the same way as the transition function. We focus hereafter on the general problem of predicting future states. 

Hidden variables $M^1,M^2,\dots, M^m$ are unknown {\it a priori} and need to be inferred from the observable entities. 
Examples of hidden variables $M^i$ include past events, or actions performed on the objects. In the previous example, a door can be unlocked or locked at a given time $t$, and the state of the door cannot be easily inferred from vision alone. The state of the door is then a hidden variable $M_t^i$ of state $S^t$. Its existence can be inferred by discovering the causal link between the act of inserting and turning a key in a door lock and the subsequent ability to open the door later in the future, after executing several possibly unrelated actions.
Transition function $T$ is represented as a {\it Dynamic Bayesian Network} (DBN). We denote by $pa(X_t)$ the list of parents of variable $X$ at time $t$. Thus, function $T$ is defined as {\small $T(S_{t+1}|S_{t},A_{t}) = \Big(\prod_{i=1}^{n} P(O_{t+1}^{i}|pa(O^{i}_{t+1}),A_t)\Big) \Big( \prod_{i=1}^{m} P(M_{t+1}^{i}|pa(M^{i}_{t+1}),A_t)\Big),$}
wherein the values of $O^{i}_{t+1}$ and $M^{i}_{t+1}$ are contained in $S_t$. 

\begin{figure}[t]
\vspace{0.2cm}
  \centering
 {
    \includegraphics[width=0.46\textwidth]{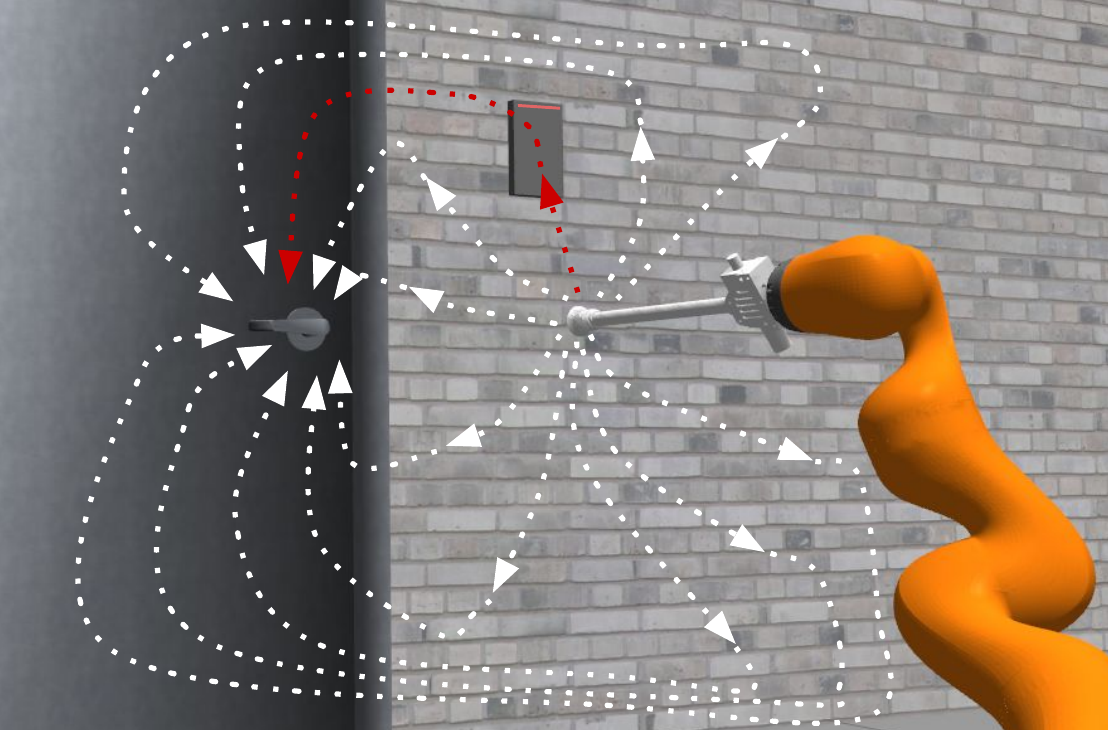}
 }
 \vspace{-0.3cm}
\caption{Learning to open a door. The robot needs to first move its end-effector to the region surrounding the card reader to unlock the door, before attempting to open the door. The causal link between the two events is time-delayed and hidden in the middle of all possible random trajectories that are generated in the exploration phase. This work proposes an algorithm for learning such models.}
 \label{fig:door_example}
 \vspace{-1.1cm}
\end{figure}

\section{Proposed Approach}

The proposed algorithm receives a set of sampled data trajectories $\mathcal{D} = \{(z^l_0,a^l_0, \dots,z^l_h,a^l_h)\}_{l=1}^{L}$, collected by executing uniformly distributed random actions $\{a^l_t\}$ for exploration. The algorithm returns a set of memory variables $\{M^i\}_{i=1}^{m}$, and parents of each variable in $\{M^i\}_{i=1}^{m} \cup \{O^i\}_{i=1}^{n}$.
We initially assume that there are no hidden variables, and incrementally create new ones only when needed to explain stochasticity of observations $\{z^l_t\}$. 
Algorithm~\ref{algo:causalLearning} summarizes the main steps of this process. 
\subsection{Identifying highly stochastic variables}
We start by setting $k$, the counter of hidden variables, to $0$, and initialize the list of parents $pa(O_{t+1}^{i})$ for each object-attribute $O^i$ with $\{O_t^1,\dots,O_t^n,A_t\}$. In other words, all attributes of objects in a given state are considered as relevant for predicting the next state. Next, factors $P(O_{t}^{i}|pa(O_t^{i}))$ of the transition function are estimated from sampled data trajectories $\mathcal{D} = \{(z^l_0,a^l_0, \dots,z^l_h,a^l_h)\}_{l=1}^{L}$. This can be achieved through any of the many existing density estimation techniques, such as frequency counts of discretized variables or the {\it Kernel Density Estimation} (KDE), as follows,
$$
P(pa(O^{i})) = \frac{\alpha w}{hL}\sum_{t=1}^{h} \sum_{l=1}^{L}\exp \Big( - w \sum_{X\in pa(O^{i})}\| X - x_t^{i,l} \|_2\Big),
$$
\begin{eqnarray}
P(O^{i},pa(O^{i})) = \frac{\alpha w}{hL}\sum_{t=1}^{h} \sum_{l=1}^{L}\exp \Big( - w \| O^{i} -o_t^{i,l} \|_2 \nonumber \\ - w \sum_{X\in pa(O^{i})}\| X - x_t^{i,l} \|_2\Big),
\label{jointKDE}
\end{eqnarray}

wherein $o_t^{i,l}$ (resp. $x_t^{i,l}$) is the observed value of variable $O^i$ (resp. $X^{i}$) in trajectory $l$ at time $t$.
The algorithm maintains a list that contains state variables that need further explanation, i.e., variables with conditional probability distributions that have an entropy above a predefined threshold $\epsilon$. For example, in the painting experiments, the variable that corresponds to the appearance of paint on the canvas is highly stochastic, even when conditioned on the pose of the brush with respect to the canvas and the executed motion. Conditional entropy is defined as,
\begin{align}
\mathbf H(X|Y_1,\dots Y_m) = & -\sum_{x\in \mathcal X} \sum_{y_1\in \mathcal Y} \dots
\sum_{y_m\in \mathcal Y} \big(  p(x,y_1,\dots,y_m)\nonumber \\ 
& \hspace{1.7cm} \log \frac{p(x,y_1,\dots,y_m)}{p(y_1,\dots,y_m)} \big),
\label{conditionalEntropy}
\end{align}

The algorithm is safeguarded against entering infinite loops by upper bounding the number of parents per variable by $max$\_$var $. The algorithm processes the variables, one at a time, until the open list is empty.

\subsection{Searching for causes of uncertainty}
The next step of the proposed algorithm consists in searching in the history of executed actions and observed object-attributes for events that can explain the stochasticity of each one of the identified highly stochastic variables, denoted by $X$ (line 9). To this end, the algorithm iterates through all attributes $O^i$ of objects (lines 12-23). For each attribute $O^i$ of each object present in the scene, including the robot's end-effector, we search in the space $\mathcal O^i$ of all possible values  for the region that brings the maximum information gain to the distribution of variable $X$. The region is defined as hyper-ball, with a center $c_i\in\mathcal O^i$ and a radius $r_i\in \mathbb R$. 
For example, $c_i$ could be a 6-dimensional point in space that corresponds to the pose of the robot's end-effector (or the manipulated object), and $r_i$ is a tolerance threshold. 
Searching in the entire space $\mathcal O^i$ is computationally inefficient. Therefore, the proposed algorithm concentrates the search in the regions of space covered in the exploration data $\mathcal D$. We thus start by estimating the distribution of $O^i$ as,
\begin{eqnarray}
P(O^{i}) = \frac{\alpha w}{hL}\sum_{t=1}^{h} \sum_{l=1}^{L}\exp \Big( - w \| O^{i} -o_t^{i,l} \|_2 \Big).
\label{marginalKDE}
\end{eqnarray}
The center of the hyper-sphere is sampled from $P(O^{i})$, while its radius $r_i$ is sampled uniformly in the interval $[0,radius(\mathcal O^i)]$. 

The center and radius of the sampled hyper-sphere are optimized in the next step (lines 16-19) by following the gradient of the information gain that occurs from adding the event of visiting the hyper-sphere into the list of parents of variable $X$. The information gain is defined as, 
\begin{eqnarray*}
IG\big((X|pa(X)),Y\big) = \mathbf H(X|pa(X)) - \mathbf H(X|pa(X), Y).
\label{infoGain}
\end{eqnarray*}

Therefore, selecting a new parent $Y$ that maximizes $IG\big((X|pa(X)),Y\big)$ corresponds to selecting a parent that minimizes $\mathbf H(X|pa(X), Y)$. In our case, $Y$ corresponds to the event of visiting $\texttt{Ball}(c_i,r_i)$ at some point in the past. Thus, we first define $\hat{\mathbf H}$, a form of $\mathbf H$ that is parameterized by $c_i$ and $r_i$, then compute the gradients of $\hat{\mathbf H}\big(X|pa(X), O^i\in \texttt{Ball}(c_i,r_i)\big)$ with respect to both $c_i$ and $r_i$, and use them to update both parameters. 

Boolean events $O^i\in \texttt{Ball}(c_i,r_i)$ are discrete and non-differentiable with respect to $c_i,r_i$. We thus employ a continuous relaxation where $P\big(X,pa(X), O^i\in \texttt{Ball}(c_i,r_i)\big)$ is defined as
\begin{eqnarray*}
\frac{\alpha w}{hL}\sum_{l=1}^{L} \sum_{t=1}^{h} \exp \big( - w \| X -x_t^{l} \|_2  - w \sum_{Y\in pa(X)}\| Y - y_t^{i,l} \|_2 \big) \\
 \exp \big( - w \min_{t'\in\{1,\dots,t-1\}}\|o_{t'}^i - c_i\|_2 - r_i\big) .
\label{relaxedProba}
\end{eqnarray*}
 $\hat{\mathbf H}\big(X|pa(X), O^i\in \texttt{Ball}(c_i,r_i)\big)$ is then defined by using $P\big(X,pa(X), O^i\in \texttt{Ball}(c_i,r_i)\big)$ in Equation~\ref{conditionalEntropy}. Although the \texttt{min} operator in this definition is not differentiable, the gradient of $\hat{\mathbf H}$
 can still be computed analytically at any point $(c_i,r_i)$ using well-known derivatives of exponential functions, except on small boundaries in the space $\mathcal O^i$, which are the boundaries of the {\it Voronoi} regions generated by the data points $\{o_{t}^i\}$. 

\subsection{Creating memory units}
After identifying key event $O^i\in \texttt{Ball}(c_i,r_i)$ and optimizing $(c_i,r_i)$ through gradient-descent, we create a binary memory variable $M^{(k)}$ associated with it. Therefore, $M^{(k)}\in\{0,1\}$. Binary hidden variables $\{M^{(k)}\}$ are used to carry pertinent information over time and preserve the Markov assumption in the transition model.
The transition function for a memory variable $M^{(k)}$ associated with object-attribute $O^i$ is defined as follows.

\vspace{-0.3cm}
{\small
\begin{align*}
        &P\big(M^{(k)}_{0} = 0 \big) = 1, \\
        &pa(M^{(k)}_{t+1})= \{ M^{(k)}_{t}, O^i_t\},\\
        &P\big(M^{(k)}_{t+1} = 1 | pa(M^{(k)}_{t+1})\big) = 1 \textrm{ if } \big( M^{(k)}_{t} = 1 \lor \|o^i_t - c_i\|_2\leq r_i \big),\\
        &P\big(M^{(k)}_{t+1} = 0 | pa(M^{(k)}_{t+1})\big) = 1  \textrm{ if } \big( M^{(k)}_{t} = 0 \land \|o^i_t - c_i\|_2 > r_i \big).
\end{align*}
}%

Therefore, memory unit $M^i_{t+1}$ is deterministic. It is initialized with $0$, and it preserves its value over time until the event $\|o^i_t - c_i\|_2\leq r_i$ occurs. This event corresponds to object $O^i$ being inside the sphere $(c_i,r_i)$ at some time $t$.

This process is repeated until the entropy of variable $X$ drops below a pre-defined threshold after identifying from the history data all hidden causes of $X$. Note also that the same object attribute $O^i$ can be associated with more than one memory unit $M^{(k)}$. This happens for example when different regions, with different centers and radii, in space $\mathcal O^i$ of object $O^i$ need to be visited in order for $X$ to take a certain value. For example, we show in the next section how a robot can learn to identify four lug nuts that need to be loosened before it can remove a tire. The locations of the nuts are not known {\it a priori} to the robot. 
\begin{figure}
\vspace{0.1cm}
   \centering
    \includegraphics[width=0.35\textwidth]{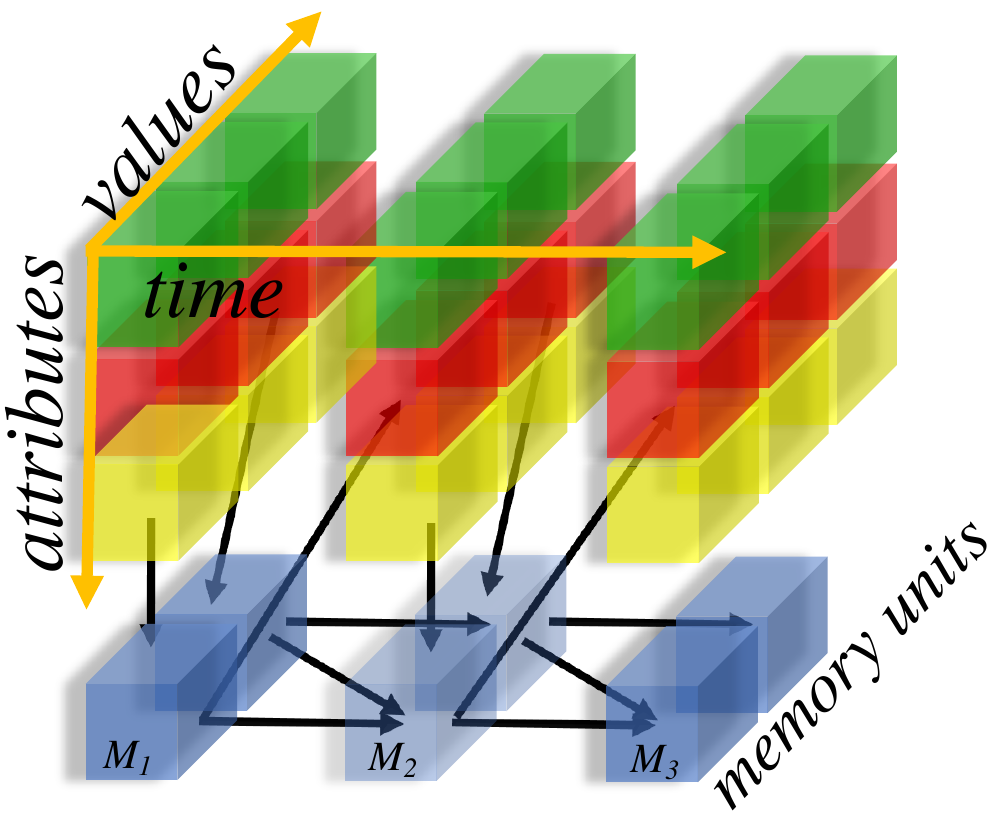}
    \caption{\footnotesize Causality graph returned by Algorithm~1.}
        \label{model}
    \vspace{-0.5cm}
\end{figure}

\newcommand\mycommfont[1]{\footnotesize\ttfamily\textcolor{blue}{#1}}
\SetCommentSty{mycommfont}
\begin{algorithm}[!t]
\footnotesize{
  \SetKwFunction{InfoGain}{InfoGain}
  \SetKwFunction{Entropy}{H}
  \SetKwFunction{ConditionalEntropy}{ConditionalEntropy}
  \SetKwFunction{CausalDirection}{CausalDirection}
\KwIn{Set of observable variables $\{O^i\}_{i=1}^{n}$, and sampled data trajectories $\mathcal{D} = \{(z^l_0,a^l_0, \dots,z^l_h,a^l_h)\}_{l=1}^{L}$;}
\KwOut{Set of memory variables $\{M^i\}_{i=1}^{m}$, and parents of each variable in $\{M^i\}_{i=1}^{m} \cup \{O^i\}_{i=1}^{n}$ \;}
$openList \leftarrow \emptyset$; $k\leftarrow 0$\;
  \tcc{\footnotesize Identifying highly stochastic variables}
\For{$i := 1; i \leq n; i\leftarrow i+1$}
{
    $\forall t\geq0: pa(O_{t+1}^{i}) \leftarrow \{O_t^1,\dots,O_t^n,A_t\}$\; 
    Estimate $P(O^{i},pa(O^{i}))$ from data $\mathcal{D}$ using Equation~\ref{jointKDE}\;
    Estimate $P(pa(O^{i}))$ from data $\mathcal{D}$ using Equation~\ref{jointKDE}\;
    Compute conditional entropy $\mathbf H(O^i|pa(O^i))$ from $P(pa(O^{i}))$ and $P(O^{i},pa(O^{i}))$ by using Equation~\ref{conditionalEntropy}\;
    \lIf{$\big(\mathbf H(O^i|pa(O^i)) > \epsilon\big)$}
    {
        $openList.push(O^{i})$
    }
}
\While{$openList \neq \emptyset$}
{
$X\leftarrow openList.pop()$\;
\tcc{\footnotesize Searching for causes of uncertainty}
    \Repeat{$\big(\mathbf H(X|pa(X))< \epsilon\big)$ $\vee$ $\big( | pa(X) | \geq $max$\_$var $\big)$}
    {
        $max\_gain\leftarrow 0$\;
        \ForEach{$i \in \{1,\dots,n\}$}
        {
            Sample an initial center-point $c_i \sim P(O^i)$\;
            Sample an initial radius $r_i \sim Uniform \big(0,radius (\mathcal O^i)\big)$\;
            \Repeat{$\Big((\|\nabla_{c_i}\hat{\mathbf H}  \|_2 < \epsilon_1)$ $\land (\|\nabla_{r_i}\hat{\mathbf H} \|_2 < \epsilon_2)\Big)$}
            {
                $\nabla_{c_i}\hat{\mathbf H} = \frac{\partial}{\partial c_i} \hat{\mathbf H}\big(X|pa(X),O^i\in \texttt{Ball}(c_i,r_i)\big)$\;
                $c_i \leftarrow c_i - \alpha_{center} \nabla_{c_i}\hat{\mathbf H}$\;
                $\nabla_{r_i}\hat{\mathbf H} = \frac{\partial}{\partial r_i} \hat{\mathbf H}\big(X|pa(X),O^i\in \texttt{Ball}(c_i,r_i)\big)$\;
                $r_i \leftarrow r_i - \alpha_{radius} \nabla_{r_i}\hat{\mathbf H}$\;
            }
            \If{$\hat{\mathbf H}\big(X|pa(X),O^i\in \texttt{Ball}(c_i,r_i)\big) - \mathbf H(X|pa(X)) > max\_gain$}
            {
                $max\_gain \leftarrow \hat{\mathbf H}\big(X|pa(X),O^i\in \texttt{Ball}(c_i,r_i)\big) - \mathbf H(X|pa(X))$\;
                $i_{max} \leftarrow i$\;
            }
        }
          \tcc{\footnotesize Adding hidden variables}
        Create a memory unit $M^{(k)}$ associated with the event $O^{i_{max}} \in \texttt{Ball} (c_{i_{max}}, r_{i_{max}})$\;
        $pa(X)\leftarrow pa(X) \cup \{M^{(k)}\}$\;   $k\leftarrow k+1$\; 
    }
}
  \DontPrintSemicolon}
\caption{Greedy Causal Graph Construction}
\label{algo:causalLearning}
\end{algorithm}

\section{Experiments}
\begin{figure}
  \centering
       \subfigure[{ Learning to paint in simulation}]
 {
    \includegraphics[width=0.45\textwidth]{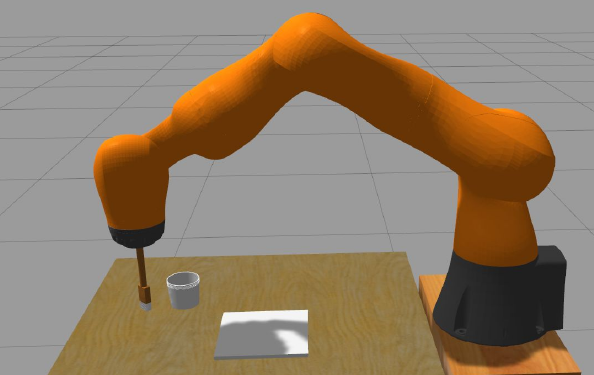}
 }
        \subfigure[{ Evaluating the learned policy with a real robot}]
 {
    \includegraphics[width=0.45\textwidth]{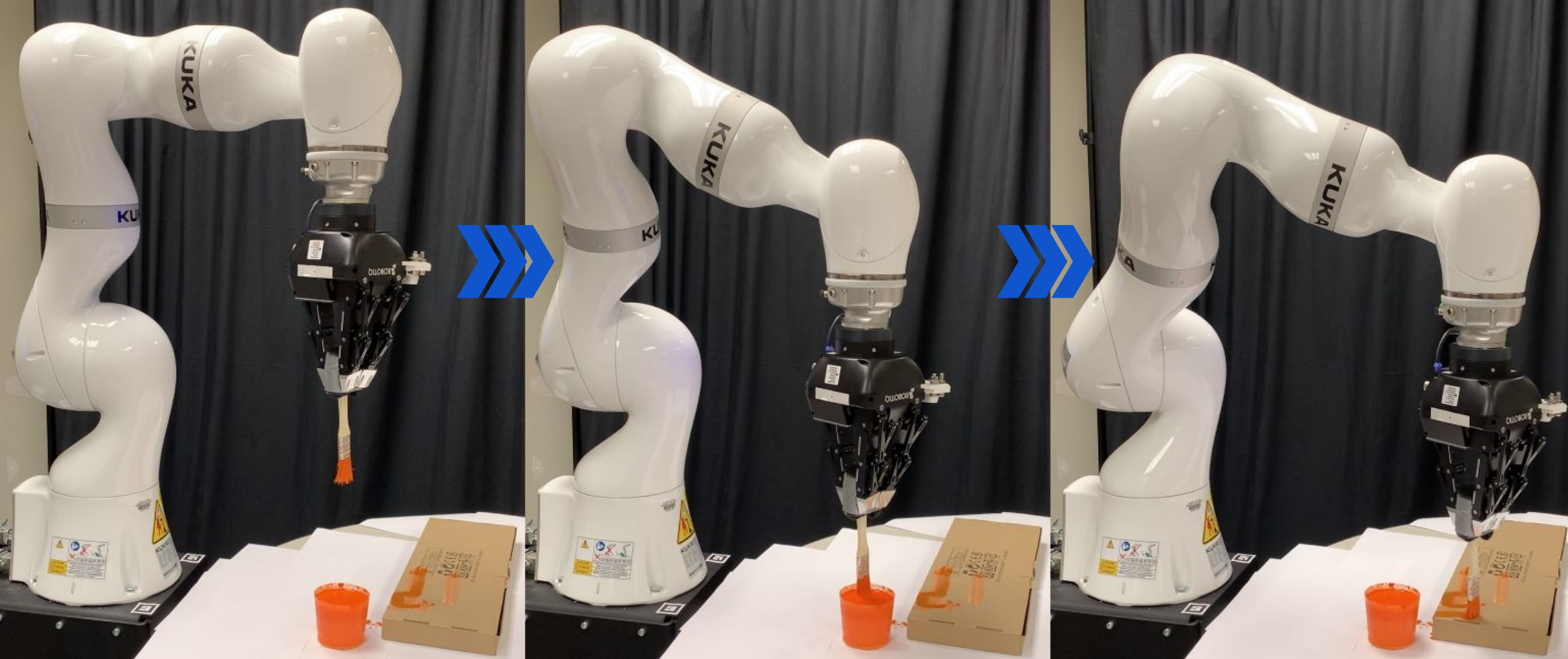}
    }
        \subfigure[{ Learning to change a tire}]
 {
    \includegraphics[width=0.45\textwidth]{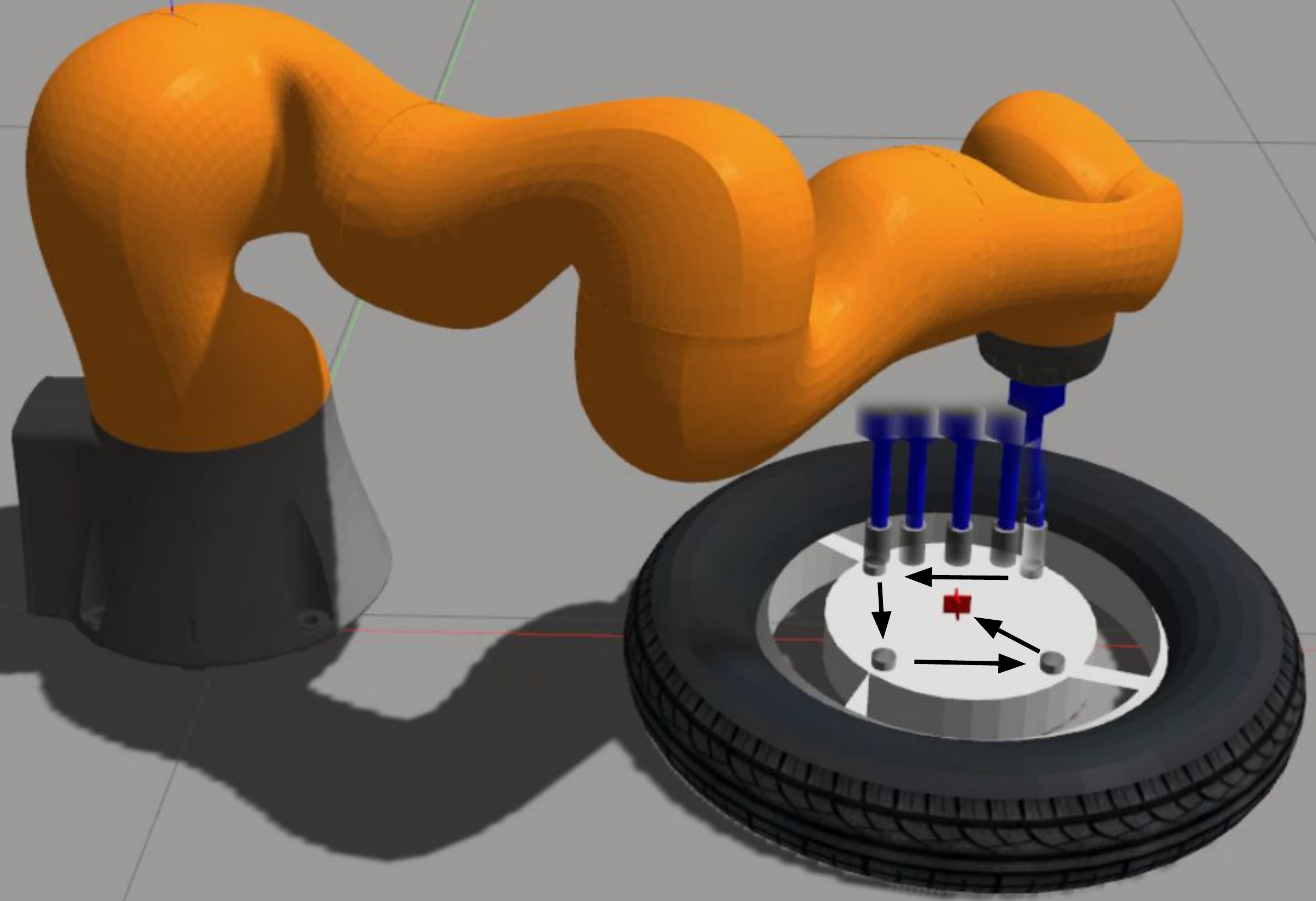}
 }
 \caption{ Tasks considered in the experiments}
 \label{fig:setups}
\end{figure}

\begin{figure}
  \centering
    \includegraphics[width=0.52\textwidth]{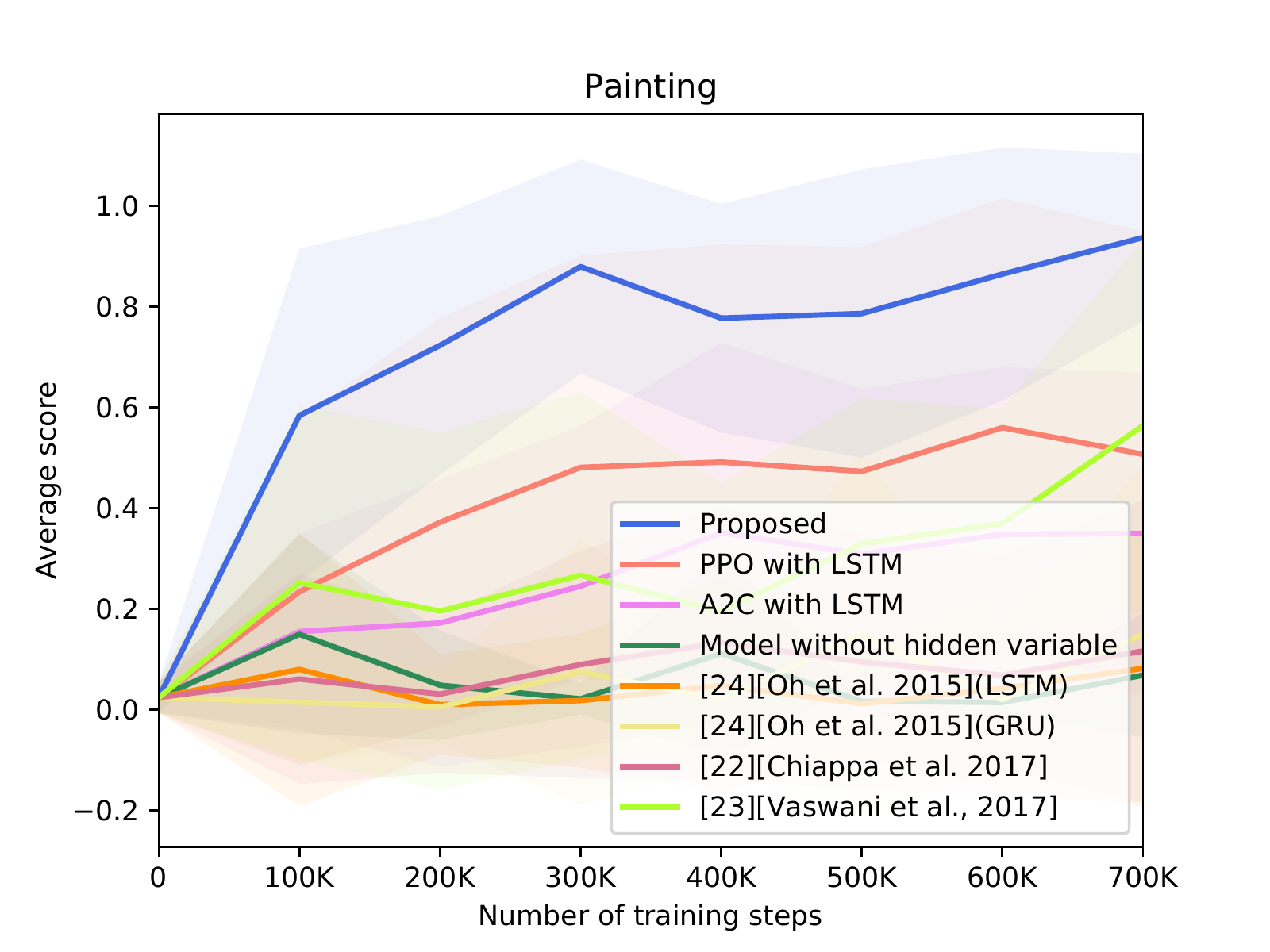}
    \includegraphics[width=0.52\textwidth]{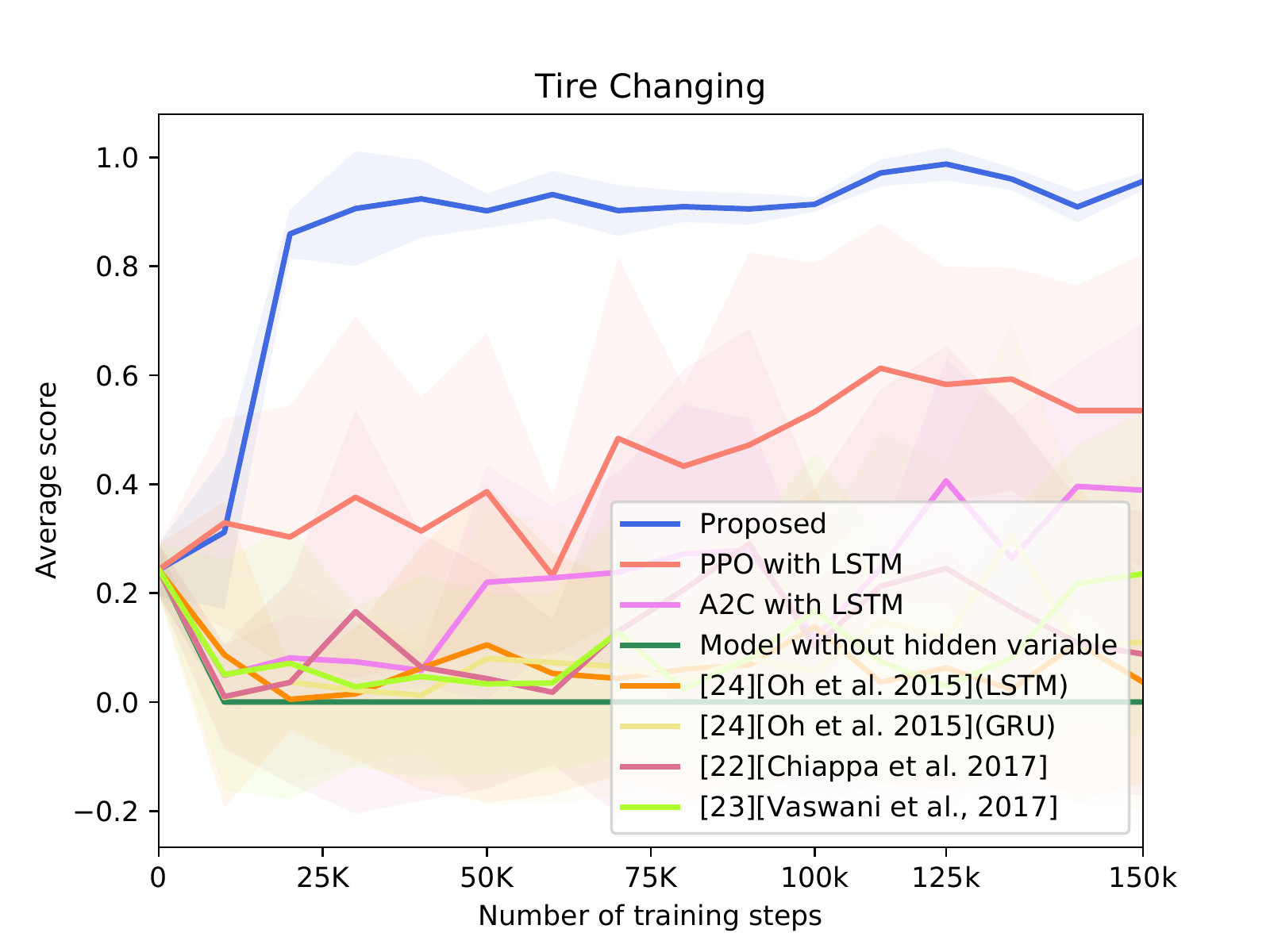}
\caption{ Average reward per test episode as a function of the number of time-steps in the training data. {\it Stacking History} refers to a variant of our architecture where the hidden variables are replaced with a long history of actions, observations and rewards.}
        \label{results}
\end{figure}

The proposed approach is evaluated on two tasks. These two are robotic experiments using the {\it Gazebo} simulator, we also deploy the proposed approach on a real robot for the painting task. 

A video of the experiments is attached as a supplementary material and uploaded to~{\textcolor{blue}{\url{https://bit.ly/2TeGlzm}}.

\subsection{Robotic Experiments}

\subsubsection{Painting}
We formulate the painting problem illustrated in Figure~\ref{fig:setups} as a POMDP, where the state space corresponds to the position of the paintbrush attached to the robot's end-effector, and a hidden binary variable that indicates if the paintbrush is loaded. It is important to note that the robot can observe only the position of the brush, and not if it is loaded or unloaded. 
The action space corresponds to movements of the end-effector in six directions. 
We use the {\it Gazebo} simulator with the {\it MoveIt!} path planner to move the end-effector between two adjacent cells. The length of each episode is $100$ time-steps.
When the brush goes inside the paint bucket, the binary variable switches from {\it false} to {\it true} and remains so until the end of the episode. This simple transition is difficult to learn because the binary switch variable can never be observed, there is no immediate evidence related to it, and the robot is even unaware of its existence a priori. If the paintbrush was loaded, the robot receives a reward of $+1$ when the brush touches the canvas. In all other cases, the received reward is $0$. The reward function is also unknown and needs to be learned from the observed trajectories. Given that the data is collected with a random policy, the time difference between dipping the brush in the bucket and touching the canvas can be arbitrarily long. 

The robot receives data sequences $\{(z_t,a_t,R_t)\}_{t=1}^{100}$, where $z_t$ is a 3D position of the brush, $a_t$ is a moving direction, and $R_t$ is an immediate reward. No other information is provided. The robot is then tasked with learning to paint.

Using the proposed algorithm, the robot learned a state transition model and a reward function from data. The learned model was then used by the value iteration algorithm to return a policy. The state transition function here is trivial, but the reward function is less trivial and involves a hidden variable related to the loading status of the brush.
The robot first noticed that the received rewards are seemingly stochastic and searched for past events that might be behind the variance in the received rewards. Searching for such events is challenging because the data is simply a collection of Brownian-like motions of the end-effector gathered using a uniform exploratory policy. Nevertheless, the algorithm succeeds in  localizing a region of state space that must be visited by the brush before pressing it on the canvas in order to make paint appear on the canvas. The algorithm systematically creates a memory unit associated with the event of visiting that specific region of the state space, which corresponds to dipping the brush in the bucket. 

Results reported in Figure~\ref{results} show that our algorithm converges to a nearly $100\%$ success rate. The results are averaged over $200$ test episodes and five different initial positions of the paint bucket and the canvas. 

\renewcommand{\tabcolsep}{1pt}
\begin{table}
\begin{center}
\small
\begin{tabular}{|c|c|c|c|c|c|}
     \hline
      Task & Proposed  & LSTM & GRU & ~\cite{chiappa2017recurrent} & ~\cite{vaswani2017attention} \\
     \hline
     Painting Reward  $0$ & $0.99 / 0.99 $   & $1 / 0.99$ &  $1 / 0.99$ & $1 / 0.99$ & $0.99 / 0.99$\\ 
     \hline
     Painting Reward $1$ & $0.90 / 0.96 $  & $0 / 0$ & $0 / 0$ & $0 / 0$ & $0.08 / 0.152$\\
     \hline
     \hline
     Tire Reward $0$ & $0.99 / 0.99 $ & $1 / 0.99$ &  $1 / 0.99$ & $1 / 0.99$ & $1 / 0.99$\\ 
     \hline
     Tire Reward $1$ & $0.96 / 0.97$  & $0 / 0$ & $0 / 0$ & $0 / 0$ & $0 / 0$\\
     \hline
\end{tabular}
\caption{\small Recall / Precision in predicting the two values of rewards. LSTM and GRU are based on~\cite{OhGLLS15}.}
\label{confusion}
\end{center}
\end{table}
We also compared the proposed method to the state-of-the-art model-free RL algorithm Proximal Policy Optimization (PPO), with an LSTM unit. PPO is implemented with a neural network that receives current position as input. The input layer is followed by fully a connected layer with $128$ and $256$ units, followed by an LSTM layer with size of $256$. The learning rate is $10^{-4}$, batch size is $4$, the number of epoch is $2$, discount factor is set to $0.99$ and the number of workers is $8$. Despite performing significantly better than a version of our method that does not search for the hidden variables, PPO underperformed in this task. Another popular model-free RL algorithm A2C\cite{pmlr-v48-mniha16} is implemented with the same setting as PPO. Its learning curve shows a similar pattern as PPO but it increases more slowly. We also compared our approach to a recent model-based RL approach for POMDPs~\cite{OhGLLS15}, where LSTM is used in one version and GRUs are used in another version. Both versions failed to learn to accurately predict the reward function using the same data used by our approach, as shown in Table~\ref{confusion}. The model learned by~\cite{OhGLLS15} was used for planning in a Monte Carlo tree search where the groudtruth optimal path was always provided intentionally  within the set of sampled paths. But the optimal path was not selected by the model and the obtained policies were suboptimal, as shown in Figure~\ref{results}. A more recent variant of~\cite{OhGLLS15},  another model-based method \cite{chiappa2017recurrent} is evaluated. But it also does not manage to find the optimal path. Attention model \cite{vaswani2017attention} is designed to focus on parts of sequences, so we evaluate whether it can find the target region in this task. Applying the same planning procedure as above, its performance is better than the other two model-based baselines, but it is still worse than the proposed approach. So even though its performance may keep increasing as more data is available, it is clearly less efficient than the proposed approach.

\subsubsection{Tire Removal}
The painting experiments involve only one hidden variable. To test the proposed algorithm on problems with more variables, we designed a second task in Gazebo where the robot is tasked with removing a tire. We assume that the robot is already equipped with an automatic drill on its end-effector, and the task consists in placing the drill on the lug nuts to loosen them before moving to the center of the wheel to take it off. The problem is formulated in the same way as in the painting task, except that the hidden variables now correspond to the status of each of four lug nut (tight vs. loose). The wheel can be taken off only when the end-effector is placed at the center of the wheel, after placing it on four specific points corresponding to the lug nuts. A reward of $1$ is given when the task is successfully finished, all other states have a reward of $0$. Results are averaged over five different positions of the nut lugs.

Results in Figure~\ref{results} and Table~\ref{confusion} confirm that the proposed approach discovers the causal link between visiting four specific regions of the state space and receiving a positive reward at the end of the episode. As the number of hidden variables increases, baseline algorithms easily get trapped in local minima. The resulting learned policies tend to go to the wheel's center after only loosening one or two lug nuts, which causes their performance to drop. Because of the clipped loss, PPO agents have limited policy changes at each update. Although it cannot improve much from the initial random policy, PPO still avoids getting trapped in a local minimum policy. A2C learns a similar policy in the beginning, but gradually improves later. In all four model-based baselines, \cite{vaswani2017attention} outperforms others in the painting experiment but it cannot show a similar advantage in this second task.  Simultaneously keeping track of multiple regions may not be trivial for current attention models.

\section{Conclusion}
Model-based RL algorithms are known to be data-efficient alternatives to model-free ones, but also to be highly sensitive to modeling inaccuracies. Such inaccuracies are often due in robotics to the partially observable nature of states that makes predicted observations and rewards  stochastic. In this work, we presented a new algorithm that  searches for  past events that can be the hidden causes behind the perceived noise in the future. The proposed model systematically builds a memory of these events and the times of their occurrences. The proposed approach searches for confounding hidden variables that once identified, can help reduce the noise of the predictions. Future works include deploying tire removal experiments on a real robot, as well as extending this framework to learning by imitation.
\section*{Acknowledgement}
This work was supported by NSF awards 1734492, 1723869 and 1846043.

\bibliographystyle{IEEEtran}
\bibliography{iros20}
\end{document}